\documentclass[11pt]{article}

\usepackage[preprint]{acl}

\usepackage{times}
\usepackage{latexsym}
\usepackage{amsmath}
\usepackage{amsfonts}
\usepackage{booktabs}
\usepackage{multirow}
\usepackage{tikz}
\usepackage{graphicx}
\usepackage{subcaption}
\usepackage{fvextra}
\usepackage[table,xcdraw]{xcolor}
\usepackage[T1]{fontenc}
\usepackage[utf8]{inputenc}
\usepackage{microtype}
\usepackage{inconsolata}
\usepackage{graphicx}
\usepackage{geometry}
\usepackage{ulem} % for underlining

\title{Closing the Modality Reasoning Gap for Speech Large Language Models}

\renewcommand{\thefootnote}{\fnsymbol{footnote}}

\author{
 \textbf{Chaoren Wang\textsuperscript{1}}\thanks{Work done during an MSRA internship.},
 \textbf{Heng Lu\textsuperscript{2}},
 \textbf{Xueyao Zhang\textsuperscript{1}},
 \textbf{Shujie Liu\textsuperscript{2}\footnotemark[2]},
\\
 \textbf{Yan Lu\textsuperscript{2}},
 \textbf{Jinyu Li\textsuperscript{2}},
 \textbf{Zhizheng Wu\textsuperscript{1}\footnotemark[2]}
\\
 \textsuperscript{1}The Chinese University of Hong Kong, Shenzhen, \\
 \textsuperscript{2}Microsoft Corporation
}

\def\modelname{TARS}

\begin{document}
\maketitle
\footnotetext[2]{Corresponding authors.}

\renewcommand{\thefootnote}{\arabic{footnote}}

\begin{abstract}
Although {Speech Large Language Models} have achieved notable progress, a substantial modality reasoning gap remains: their reasoning performance on speech inputs is markedly weaker than on text. 
This gap could be associated with representational drift across Transformer layers and behavior deviations in long-chain reasoning.
To address this issue, we introduce \textbf{TARS}, a reinforcement‑learning framework that aligns text‑conditioned and speech‑conditioned trajectories through an asymmetric reward design.
The framework employs two dense and complementary signals: representation alignment, which measures layer-wise hidden-state similarity between speech- and text-conditioned trajectories, and behavior alignment, which evaluates semantic consistency between generated outputs and reference text completions. 
Experiments on challenging reasoning benchmarks, including MMSU and OBQA, show that our approach significantly narrows the modality reasoning gap and achieves state-of-the-art performance among 7B-scale Speech LLMs.
\end{abstract}

\section{Introduction}

%brief introduce the speech llm
%why speech llm has modality reasoning gap  --> modality alignment
% https://neurips.cc/virtual/2025/loc/san-diego/136685
% multimodal tax, capacity split (split brain with finite compute), distribution shift (multimodal data dominates the loss landscape, text reasoning ability degraded)

Recent advances in Speech Large Language Models (Speech LLMs) enable a unified framework for spoken language processing tasks such as automatic speech recognition (ASR), speech translation, and speech QA.
Most Speech LLMs follow a three-stage architecture, consisting of a pre-trained speech encoder, lightweight adapters, and a decoder-only text LLM~\cite{speech_survey2025,speech_survey2025_weng}. 
The encoder transforms raw speech into high‑resolution acoustic representations, which are then projected into the text‑embedding space through learned adapters, enabling the downstream LLM to process speech inputs using its text‑native reasoning stack.
Through shared representations across modalities, this architecture allows speech inputs to leverage the generation and reasoning capabilities of text-based LLMs.

However, Speech LLMs exhibit a persistent and critical challenge: the modality reasoning gap, denoting a substantial decline in reasoning performance on speech inputs compared to text, as evidenced by empirical analyses~\cite{emnlp} on VoiceBench and SpeechMMLU benchmarks~\cite{voicebench,mimoaudio}.

%% draft v2: input, process, outcome three level problem + need for RL (on-policy)
To close modality reasoning gap, prior works have primarily focused on input-side fusion and output-side supervision.
Input-level modality fusion methods aim to reduce the discrepancy between speech representations and text embeddings at input stage, by freezing the LLM backbone and training lightweight adapters~\cite{alignchat,desta,xu2025optimal}. 
As speech naturally contains paralinguistic cues absent in text, strict input equivalence may not be desirable. However, for complex reasoning tasks, the underlying logical progression should remain invariant regardless of the input modality~\cite{mousavi2025alas}. Relying solely on inputs can cause subtle discrepancies to propagate and amplify through Transformer layers, leading to significant representational drift.
Another line of work provides output-level supervision. They focus on knowledge distillation or prompt-switching training to encourage speech-conditioned generations to mimic text-conditioned behaviors~\cite{kd,kimi}. 
However, these methods enforce strict token-level supervision in {an} off-policy manner. Since the speech-conditioned distribution differs from the text one, forcing the model to generate exact text-conditioned tokens targets an unreachable objective. Furthermore, this rigid supervision suffers from exposure bias: a single token error shifts the model into an unsupervised state, causing it to fall into behavioral divergence.

% introduction of our method
%% draft v2: intuition motivation
To address these limitations, we introduce \textbf{TARS} (\textbf{T}rajectory \textbf{A}lignment for \textbf{R}easoning in \textbf{S}peech), which combines representation alignment to mitigate drift and behavior alignment with more flexible objective. By leveraging on-policy exploration, our method enables the model to mitigate exposure bias and maintain reasoning stability, preventing the model from drifting into out-of-distribution states.
Specifically, representation alignment is computed from the cosine similarity of layer‑wise hidden states, providing coarse‑grained representation‑level feedback. Complementarily, behavior alignment is derived from the semantic consistency of the final outputs, providing token‑level but comparatively sparser feedback. These signals effectively steer the speech modality toward the text reasoning trajectory.

Under an asymmetric reward design with Group Relative Policy Optimization (GRPO)~\cite{grpo}, we jointly optimize task accuracy and these two alignment rewards.
Even when all samples in the generated group exhibit zero task accuracy, a common outcome for speech‑conditioned reasoning given its greater difficulty than text‑based inference, the alignment signals remain informative, enabling direct trajectory alignment between speech and text.
Empirical results on complex reasoning benchmarks, such as MMSU and OBQA~{\cite{voicebench}}, demonstrate that our method outperforms existing baselines, achieving state-of-the-art performance among 7B models.
% our contribution
The contributions of our work are summarized as follows:
\begin{itemize}
% method level:
\item We propose an on-policy reinforcement-learning framework for {trajectory alignment} that aligns speech-conditioned reasoning trajectory with its text-conditioned counterpart, closing the modality reasoning gap without architectural modifications.

% reward design
\item We introduce an asymmetric dense reward with two complementary alignment signals: {representation alignment} that reduces layer-wise hidden-state drift, and {behavior alignment} that enforces semantic consistency.

\item Our method achieves the state‑of‑the‑art performance on reasoning benchmarks (MMSU, OBQA) among 7B‑scale models. We release datasets and code to facilitate reproducibility.\footnote{\url{https://github.com/AmphionTeam/TARS}}
\end{itemize}

\section{Related Works}

\paragraph{Speech LLMs.}
Speech LLMs have progressed from cascaded pipelines (ASR + text LLM + optional TTS) to end-to-end architectures that couple speech perception with LLM-style generation, enabling spoken dialogue and spoken QA while better leveraging paralinguistic cues beyond transcripts~\cite{speech_survey2025,speech_survey2025_weng}.
A dominant design follows a three-stage paradigm: a pretrained speech encoder extracts acoustic features, which are mapped into the text embedding space via lightweight projectors to condition a decoder-only LLM, preserving text-pretrained reasoning priors while extending to speech tasks~\cite{salmonn,qwen-audio,listen_think_understand}.
Recent open and proprietary omni systems further target low-latency and multi-modal interaction.
Open-weight examples such as Qwen2.5-Omni and Qwen3-Omni integrate unified perception and generation, and introduce modality-specialized capacity (e.g., MoE routing) to improve scalability~\cite{xu2025qwen2,qwen3}.
Audio-centric models like Kimi-Audio and MiniCPM-o emphasize practical voice interaction and general audio understanding~\cite{kimi,yao2024minicpm}.

\paragraph{Modality Alignment and Reasoning Gap.}
Despite unified architectures, a performance disparity between speech and text modalities persists, termed the modality reasoning gap. Empirical studies~\cite{voicebench,mimoaudio,mousavi2025alas} and representational analyses~\cite{emnlp} reveal that speech-conditioned hidden states often drift from their text counterparts, leading to degraded reasoning.
Existing efforts to bridge this gap generally fall into two categories: 
\textbf{(1) Architectural Adaptations.}
To preserve text-based capabilities, a prominent line of work adopts a frozen-backbone strategy. Methods such as AlignChat~\cite{alignchat}, DeSTA~\cite{desta}, OTReg~\cite{xu2025optimal}, and MTBI~\cite{xie2025enhancing} keep Large Language Model (LLM) parameters fixed, focusing exclusively on refining the input-side projector. However, this yields only surface-level alignment. By freezing the backbone, the model cannot adapt to speech-specific dynamics, causing reasoning trajectories to diverge even when inputs are closely projected.
\textbf{(2) Supervised Alignment Strategies.}
Other works employ static matching during pretraining or supervised fine-tuning. Early methods explored manifold mixup~\cite{stemm} and contrastive objectives~\cite{cappellazzo2024} to align representation spaces. Recent LLM adaptations employ joint speech-text supervised fine-tuning~\cite{voicetextblender}, task-independent contrastive pretraining~\cite{zufle2025}, or interleaving tokens directly~\cite{spiritlm}. To align output behaviors, approaches like Kimi-Audio~\cite{kimi} utilize prompt-switching, while others apply cross-modal knowledge distillation~\cite{kd} or data selection~\cite{salad} to match text-based teachers. Works like SSR~\cite{tan-etal-2025-ssr} combine both, aiming to align representation and output behavior simultaneously.
However, these methods rely on off-policy supervision (e.g., cross-entropy or KL divergence) on static targets. This forces the model to mimic final answers but does not teach it how to dynamically correct its own reasoning trajectory, leading to compounding errors in complex tasks. In contrast, our reinforcement learning (RL) based framework aligns the reasoning trajectory itself via on-policy exploration.

\paragraph{Reinforcement Learning for Reasoning.}
RL has proven essential for enhancing the reasoning capabilities of LLMs beyond standard supervised fine-tuning~\cite{liu2025reinforcement}. Techniques like GRPO enable models to learn from sparse rewards and self-exploration, significantly improving performance on math and logic tasks~\cite{grpo,dapo}.
In the speech domain, however, RL application remains nascent, primarily limited to aligning paralinguistic attributes or general helpfulness rather than reasoning logic~\cite{li2025emo,liu2025enhancing}. A concurrent work, SoundMind-RL~\cite{soundmindrl}, also applies RL to Qwen2.5-Omni for speech reasoning. Despite this, they rely on sparse, rule-based rewards (format and correctness constraints) to shape outputs. 
Most importantly, standard binary rewards are sparse and insufficient for modality alignment.
Our work bridges this gap by adapting RL with dense alignment signals--leveraging the text modality as a stable reference to guide the speech reasoning trajectory, explicitly aligning latent spaces and mitigating representation drift.

\section{Method}
\label{sec:methodology}

\begin{figure*}[ht]
    \centering
    \includegraphics[width=\linewidth]{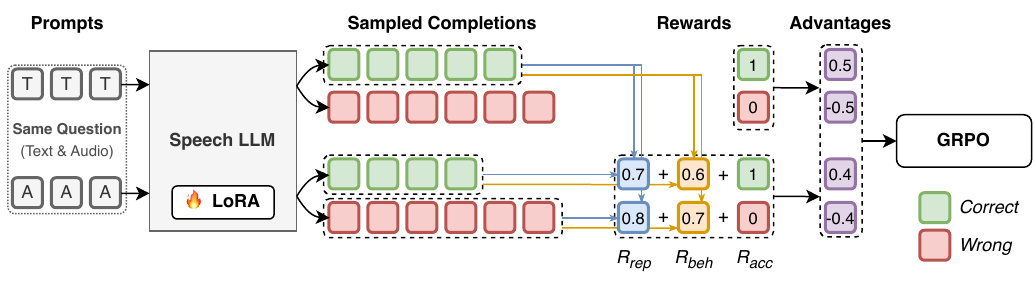}
    \caption{{Overview of our framework.} We introduce a reinforcement learning approach for trajectory alignment by optimizing an asymmetric reward function composed of representation alignment and behavior alignment.}
    \label{fig:overview}
\end{figure*}

\subsection{Problem Formulation}
\label{sec:problem_formulation}
We define a Speech LLM $\pi_{\theta}$ as a composite architecture consisting of an audio encoder, a modality projector, and a decoder-only LLM initialized from a text-pretrained LLM $\pi_{\text{base}}$. For a given query $q \in \mathcal{D}$, the model accepts either its speech representation $q_{\text{speech}}$ or text representation $q_{\text{text}}$ as input to generate a text response $y$.
Despite extensive alignment training on large-scale speech-text pairs, a significant {modality reasoning gap} persists, where the model's performance on speech inputs lags behind its text capabilities.

We quantify this gap using {Modality Recovery Rate (MRR)}. Let $y_{\text{speech}} = \pi_{\theta}(q_{\text{speech}})$ be the completion generated by the current model, and $y_{\text{text}}^{\text{base}} = \pi_{\text{base}}(q_{\text{text}})$ be the reference completion from the base model. Given a reasoning metric $\mathcal{S}$ (e.g., Question Answering (QA) accuracy), MRR measures the extent to which the Speech LLM retains the original reasoning capability:

\begin{equation}
    \text{MRR}(\pi_{\theta}) = \frac{\mathbb{E}_{q \in \mathcal{D}} [\mathcal{S}(y_{\text{speech}})]}{\mathbb{E}_{q \in \mathcal{D}} [\mathcal{S}(y_{\text{text}}^{\text{base}})]} \times 100\%.
    \label{eq:mrr}
\end{equation}
Our objective is to optimize parameters $\theta$ such that $\text{MRR} \geq 100\%$, effectively closing the gap.

% Move to the most imporant reward design
\subsection{Reward Modeling}
\label{sec:reward_design}

We propose an asymmetric reward design to align reasoning trajectories across modalities. During training, we use text-conditioned completions $y_{\text{text}}$ generated by the current policy $\pi_{\theta}$ as a moving reference. We optimize the policy on both text-conditioned and speech-conditioned completions, allowing the text branch to continue improving under base reward while providing an increasingly strong reference for aligning speech trajectories. As a result, the speech modality co-evolves with the model's improving text reasoning capability.

For a speech-conditioned completion $y_{\text{speech}}$, the total reward is defined as:
\begin{equation}
    R_{\text{total}} = R_{\text{base}} + \alpha \cdot R_{\text{rep}} + \beta \cdot R_{\text{beh}},
    \label{eq:total_reward}
\end{equation}
where $R_{\text{rep}}$ and $R_{\text{beh}}$ correspond to representation alignment and behavior alignment signals, respectively. We apply $R_{\text{total}}$ to speech-conditioned completions, while text-conditioned completions are optimized using $R_{\text{base}}$. We set $\alpha=1.0$ and $\beta=1.0$ in our experiments to simultaneously align internal representations and external behaviors.

\paragraph{Base Reward.}

Following the formulation in DeepSeek-R1~\cite{deepseekr1}, we design the base reward to optimize task accuracy and output format. This configuration serves as the {Standard GRPO} baseline in our ablation studies. It is computed as:
\begin{equation}
    R_{\text{base}} = R_{\text{acc}} + \lambda R_{\text{fmt}},
\end{equation}
where $R_{\text{acc}}\in \{0, 1\}$ indicates whether the answer extracted by xFinder~\cite{xfinder} matches the ground truth, and $R_{\text{fmt}} \in \{0, 1\}$ rewards format compliance (see Appendix~\ref{app:regex}). We set $\lambda=0.5$.

\paragraph{Representation Alignment Reward.}

To align internal representations, we compute the layer-wise geometric similarity between speech completions and text references.
Let $\mathbf{H}^{(l)} \in \mathbb{R}^{T \times d}$ denote the hidden states at layer $l$. To handle the variable sequence lengths of different modalities, we mean-pool the generated reasoning tokens (excluding $n$ prompt tokens) into a fixed-size vector $\bar{\mathbf{h}}^{(l)}$:
\begin{equation}
    \bar{\mathbf{h}}^{(l)} = \frac{1}{T-n} \sum_{t=n+1}^{T} \mathbf{H}^{(l)}_{t,:}.
\end{equation}
For each speech completion, we randomly sample a correct text completion ($R_{\text{acc}} = 1$) from the same group as a reference, computing the average cosine similarity across $L$ layers:
\begin{equation}
    R_{\text{rep}} = \frac{1}{L} \sum_{l=1}^{L} \text{CosSim}(\bar{\mathbf{h}}_{\text{speech}}^{(l)}, \bar{\mathbf{h}}_{\text{text}}^{(l)}).
\end{equation}
If no correct text completion exists, we set $R_{\text{rep}}=0$. This fallback is rare (5.9\%--10.7\% of cases), ensuring most groups provide robust gradient signals. Ultimately, this dense reward encourages the speech modality to emulate text-based internal reasoning.

% - Why not last token's hidden state? Bad performance.

\paragraph{Behavior Alignment Reward.}

To ensure behavior consistency at the output level, we employ an external embedding model $\mathcal{E}$ (e.g., \texttt{Qwen3-Embedding-0.6B}~\cite{qwen3embed}) to measures the semantic equivalence between the final generated speech completion $y_{\text{speech}}$ and text reference $y_{\text{text}}^*$:
\begin{equation}
    R_{\text{beh}} = \text{CosSim}(\mathcal{E}(y_{\text{speech}}), \mathcal{E}(y_{\text{text}}^*)).
\end{equation}
Similarly, if no correct text completion is available, we set $R_{\text{beh}}=0$.
This objective allows the model to learn from diverse valid reasoning trajectories, provided the final semantic behavior remains consistent with the teacher.

% Implement the Reward Signal using RL (GRPO)
\subsection{Reinforcement Learning Framework}
\label{sec:framework}

As illustrated in Figure~\ref{fig:overview}, we employ GRPO to optimize our proposed reward. For a given prompt $q$, we generate a group of $G$ completions, composed of equal numbers of speech-conditioned and text-conditioned completions. Following GRPO Done Right (Dr.~GRPO)~\cite{drgrpo}, we define the advantage $\hat{A}_i$ for the $i$-th completion as normalizing its reward against the group's mean. The model's parameters $\theta$ are updated using the DAPO loss~\cite{dapo}. Compared to using only the base reward, our alignment reward provides richer guidance by supplying a continuous similarity-based signal that remains effective even when task accuracy rewards are uniformly zero for speech-conditioned reasoning trajectories.

\paragraph{Modality Specific Normalization.}
A naive implementation of GRPO normalizes reward across the entire group. However, text-conditioned completions inherently achieve higher base rewards than speech-conditioned, which would cause speech-conditioned completions to consistently receive negative advantages, suppressing learning. To address this, we introduce modality-specific normalization, calculating advantages for text and speech completions in separate groups:
\begin{equation}
    \hat{A}_{i, m} = r_{i, m} - \mu_m, \quad m \in \{\text{speech}, \text{text}\},
\end{equation}
where $\mu_m$ is the mean of rewards within modality $m$. This ensures that each modality is optimized relative to its own baseline, allowing continuous improvement in modality alignment.

\section{Experiments}
\label{sec:experiments}

\begin{table*}[t!]
    \centering
    \resizebox{\textwidth}{!}{
    \begin{tabular}{l|c|cc|cc|cc|c}
        \toprule
        \multirow{2}{*}{\textbf{Model}} & \multirow{2}{*}{\textbf{Backbone}} & \multicolumn{2}{c|}{\textbf{MMSU}} & \multicolumn{2}{c|}{\textbf{OBQA}} & \multicolumn{2}{c|}{\textbf{Average}} & \multirow{2}{*}{\textbf{MRR (\%)}} \\
        \cmidrule(lr){3-4} \cmidrule(lr){5-6} \cmidrule(lr){7-8}
        & & \textbf{A} & \textbf{T} & \textbf{A} & \textbf{T} & \textbf{A} & \textbf{T} & \\
        \midrule
        \rowcolor{gray!10} \multicolumn{9}{l}{\textit{\textbf{Proprietary \& Cascaded Systems}}} \\
        GPT-4o-mini-Audio & GPT-4o-mini & 72.90 & 81.23 & 84.84 & 90.11 & 78.87 & \underline{85.67} & 92.06 \\
        ASR$^\dagger$ + Llama3.1-8B & Llama3.1-8B & 58.78 & 65.65 & 72.53 & 80.88 & 65.66 & \underline{73.27} & 89.61 \\
        ASR$^\dagger$ + Qwen2.5-7B & Qwen2.5-7B & 67.1 & 71.65 & 84.0 & 83.74 & 75.55 & \underline{77.70} & 97.23 \\
        ASR$^\dagger$ + Phi-4-7B & Phi-4-7B & 69.00 & 74.92 & 77.80 & 83.96 & 73.40 & \underline{79.44} & 92.40 \\
        \midrule
        \rowcolor{gray!10} \multicolumn{9}{l}{\textit{\textbf{Existing Baselines}}} \\
        DeSTA2.5-Audio & Llama3.1-8B$^*$ & 60.87 & 65.65 & 74.06 & 80.88 & 67.47 & 73.27 & 92.08 \\
        SALAD-7B & Qwen2.5-7B & 57.5 & 71.6 & 75.1 & 90.1 & 66.30 & 80.85 & 85.33 \\
        MiniCPM-o 2.6 & Qwen2.5-7B & 54.78 & 59.42 & 78.02 & 82.86 & 66.40 & 71.14 & 85.46 \\
        Knowledge Distillation & Qwen2.5-7B & 63.09 & 69.15 & 82.64 & 84.62 & 72.87 & 76.89 & 93.78 \\
        AlignChat & Qwen2.5-7B$^*$ & 69.65 & 71.65 & 85.49 & 83.74 & 77.57 & 77.70 & 99.83 \\
        \midrule
        \rowcolor{gray!10} \multicolumn{9}{l}{\textit{\textbf{Base \& Aligned Models}}} \\
        Qwen2.5-Omni & Qwen2.5-7B & 61.51 & 67.94 & 81.09 & 84.40 & 71.30 & 76.17 & 91.76 \\
        Phi-4-MM & Phi-4-7B & 54.81 & 72.15 & 71.65 & 84.62 & 63.23 & 78.39 & 79.59 \\
        \textbf{\modelname~(Qwen2.5-Omni)} & Qwen2.5-7B & 67.96 & 68.54 & 85.71 & 88.57 & 76.84 & 78.56 & 98.89 \\
        \textbf{\modelname~(Phi-4-MM)} & Phi-4-7B & \textbf{70.14} & \textbf{75.76} & \textbf{89.45} & \textbf{91.87} & \textbf{79.80} & \textbf{83.82} & \textbf{100.45} \\
        \bottomrule
    \end{tabular}
    }
    \caption{{Reasoning Benchmarks Results.} Accuracy (\%) on MMSU and OBQA are reported using VoiceBench evaluator for Audio (\textbf{A}) and Text (\textbf{T}) input. Underlined scores denote the values used as the denominator when computing MRR. $^\dagger$ cascaded systems; $^*$ frozen LLM backbone. {Cited results keep original precision.}}
    \label{tab:main_results}
\end{table*}

We evaluate whether our approach reduces the speech-text modality reasoning gap. Performance is reported on Audio ({A}) and Text ({T}) inputs, using {MRR} (Equation~\ref{eq:mrr}) to quantify reasoning recovery in speech modality. Our method centers on an alignment-aware {RL} framework with asymmetric rewards and modality-specific normalization to optimize reasoning.

\subsection{Experimental Setup}
\label{sec:exp_setup}

\paragraph{Foundation Models.}
We evaluate two $\sim$7B composite Speech LLMs (Section~\ref{sec:problem_formulation}): the officially released Qwen2.5-Omni and an internal Phi-4-Multimodal-Instruct (Phi-4-MM)\footnote{Same architecture as the official 5.6B release, but scaled to 7B.} built on the Phi-4-7B backbone with a lightweight speech encoder. All post-training experiments share these initial checkpoints to ensure fair comparison.

\paragraph{Training Data.}
We use the UnifiedQA training set~\cite{unifiedqa}\footnote{{Specifically, the \texttt{auxiliary\_train} split from the MMLU repository (\url{https://huggingface.co/datasets/cais/mmlu}).}} as our primary dataset. To construct paired speech-text inputs, we synthesize speech using two high-quality TTS systems: CosyVoice2~\cite{cosyvoice2} and \texttt{openaudio-s1-mini}\footnote{\url{https://huggingface.co/fishaudio/openaudio-s1-mini}}, with reference speakers sampled from the Emilia-YODAS EN subset~\cite{emilia}. After filtering the synthesized speech with a 10\% WER threshold via \texttt{whisper-medium}~\cite{whisper}, our final training set comprises 9,953 samples (203 hours).

\paragraph{Evaluation Benchmarks.}
We evaluate on two spoken multiple-choice QA benchmarks from VoiceBench~\cite{voicebench}: MMSU (3,074 examples derived from MMLU-Pro~\cite{mmlupro}) for multi-domain knowledge and reasoning, and OBQA (455 examples) for elementary science facts and commonsense reasoning. We also report {WER} on LibriSpeech~\cite{librispeech} (test-clean \& test-other) to diagnose ASR capabilitie, obtained by prompting the model to transcribe the speech and computing WER with greedy decoding.

\paragraph{Baselines.}
All post-training baselines share the same foundation models, data, and LoRA configurations. We compare against three categories (details in Appendix~\ref{app:baseline_details}): (1) \textbf{Cascaded \& Proprietary systems}: \texttt{whisper-large-v3} paired with corresponding text backbones, and GPT-4o-mini-Audio. (2) \textbf{Cross-modal alignment}: SALAD~\cite{salad}, DeSTA2.5-Audio~\cite{desta25}, AlignChat~\cite{alignchat}, and Knowledge Distillation (KD)~\cite{kd}, which explicitly align speech and text via distillation or adapter tuning. (3) \textbf{General post-training}: Supervised Fine-Tuning (SFT), Direct Preference Optimization (DPO), and Standard GRPO, which optimize for task accuracy without explicitly addressing the modality gap.

\paragraph{RL Training Protocol.}
We follow the GRPO training protocol with a group size of \(G=8\), comprising equal speech- and text-conditioned completions per prompt. During RL (Section~\ref{sec:reward_design}), text completions from the current policy serve as alignment references (teacher signals) and also participate in gradient updates. Finally, we compute advantages with modality-specific normalization to prevent consistently negative advantages for speech completions (Section~\ref{sec:framework}).

\paragraph{Implementation Details.} 
Training is conducted via ms-swift\footnote{\url{https://github.com/modelscope/ms-swift}} for Qwen2.5-Omni and HuggingFace TRL\footnote{\url{https://github.com/huggingface/trl}} for Phi-4-MM. We apply parameter-efficient fine-tuning using LoRA to all linear layers while freezing the audio encoder and projector. For RL, we use the DAPO loss estimator~\cite{dapo} (Section~\ref{sec:framework}) and sampling decoding to encourage exploration. Greedy decoding is used for deterministic evaluation, with xFinder~\cite{xfinder} extracting predicted options to reduce formatting noise. Training requires $\sim$55 hours for Qwen2.5-Omni and $\sim$35 hours for Phi-4-MM on 4$\times$A100 or 8$\times$H200 GPUs (details in Appendices~\ref{app:hyperparams} and~\ref{app:prompts}).

\subsection{Main Results}
\label{sec:exp_main_results}

Table~\ref{tab:main_results} presents the performance of our framework against a suit of baselines. As shown by the base Speech LLMs, performance under speech inputs consistently lags behind text, revealing a clear modality reasoning gap. Existing cross-modal alignment baselines can narrow this gap, yet most still fall short of full recovery with MRR < 100\%.
Our proposed framework demonstrates state-of-the-art performance among 7B models. 
For the Qwen2.5-Omni with Qwen2.5-7B as backbone, our approach achieves an average audio accuracy of \textbf{76.84\%}, substantially outperforming other end-to-end alignment methods such as SALAD (66.30\%) and MiniCPM-o (66.40\%). It also reaches an MRR of \textbf{98.89\%}. Our RL-based method proves more effective than supervised mimicry, surpassing the Knowledge Distillation (KD) baseline (72.87\%) by a large margin.
For the Phi-4-MM with Phi-4-7B as backbone, our method achieves the best performance with an accuracy of \textbf{79.80\%}, even surpassing the original text accuracy of 78.39\% and achieving the MRR = \textbf{100.45\%}. 

These results show that our method not only narrows the modality reasoning gap, but also improves text performance, from 76.17\% to 78.56\% for Qwen2.5-Omni and from 78.39\% to 83.82\% for Phi-4-MM, indicating that gains in speech are not obtained at the expense of text reasoning, instead, the knowledge learned from speech can further strengthen text‑based reasoning.
Notably, the improved text accuracy remains higher than the corresponding audio accuracy (76.84\% and 79.80\% respectively), suggesting that residual differences are likely due to imperfect speech representations and cross-modal projection noise, making text inputs a natural upper bound.

Cascaded systems are often considered strong baselines and can outperform end-to-end models.
However, our end-to-end models on Qwen2.5-Omni (76.84\%) and Phi-4-MM (79.80\%) exceed the performance of the ASR + Qwen2.5-7B pipeline (75.55\%) and ASR + Phi-4-7B (73.40\%), respectively.
This suggests that directly processing speech signals can avoid certain ASR-induced errors, leading to a more robust reasoning process.

\subsection{Effectiveness of Training Strategies}
\label{sec:rl_effectiveness}

\begin{table*}[t!]
    \centering
    \resizebox{\textwidth}{!}{
    \begin{tabular}{l|cc|cc|cc|c|c}
        \toprule
        \multirow{2}{*}{\textbf{Method}} & \multicolumn{2}{c|}{\textbf{MMSU}} & \multicolumn{2}{c|}{\textbf{OBQA}} & \multicolumn{2}{c|}{\textbf{Average}} & \multirow{2}{*}{\textbf{MRR (\%)}} & \multirow{2}{*}{\textbf{WER (\%)}} \\
        \cmidrule(lr){2-3} \cmidrule(lr){4-5} \cmidrule(lr){6-7}
        & \textbf{A} & \textbf{T} & \textbf{A} & \textbf{T} & \textbf{A} & \textbf{T} & & \\
        \midrule
        \rowcolor{gray!10} \multicolumn{9}{l}{\textit{\textbf{Inference Baseline}}} \\
        Phi-4-MM & 54.00 & 71.15 & 72.31 & 85.05 & 63.16 & 78.10 & 79.60 & \textbf{4.16} \\
        \quad + CoT Prompting & 60.77 & 70.85 & 79.34 & 86.15 & 70.06 & 78.50 & 88.29 & - \\
        \midrule
        \rowcolor{gray!10} \multicolumn{9}{l}{\textit{\textbf{Post-training Alignment}}} \\
        SFT & 63.50 & 70.85 & 81.54 & 87.47 & 72.52 & 79.16 & 91.37 & 4.18 \\
        DPO & 66.33 & 74.72 & 84.40 & 91.43 & 75.37 & 83.08 & 94.98 & 4.23 \\
        \midrule
        \rowcolor{gray!10} \multicolumn{9}{l}{\textit{\textbf{Reinforcement Learning}}} \\
        Text-only GRPO & 67.79 & \textbf{76.97} & 85.93 & 90.55 & 76.86 & \textbf{83.76} & 96.86 & 4.18 \\
        Standard GRPO ($R_{\text{base}}$) & 63.04 & 72.54 & 83.30 & 89.45 & 73.17 & 81.00 & 92.21 & 4.24 \\
        \quad + Representation ($R_{\text{rep}}$) & 66.82 & 76.09 & 84.84 & 88.35 & 75.83 & 82.22 & 95.56 & 4.18 \\
        \quad + Behavior ($R_{\text{beh}}$) & 69.55 & 76.19 & 87.91 & 90.99 & 78.73 & 83.59 & 99.22 & 4.20 \\
        \textbf{\modelname} & \textbf{69.90} & 75.47 & \textbf{89.23} & \textbf{91.65} & \textbf{79.57} & 83.56 & \textbf{100.28} & 4.20 \\
        \bottomrule
    \end{tabular}
    }
    \caption{{Analysis of Training Strategies and Reward Components.} Comparisons on the Phi-4-MM backbone using xFinder evaluator. {WER} reports the average Word Error Rate ($\downarrow$) on Librispeech.}
    \label{tab:rl_effectiveness}
\end{table*}

Table~\ref{tab:rl_effectiveness} compares different training strategies on the same backbone (Phi-4-MM), including inference-time prompting, SFT, DPO, Standard GRPO, and our method. Chain-of-Thought (CoT) prompting yields a clear gain on speech inputs, improving the average audio accuracy from 63.16\% to 70.06\% and increasing MRR from 79.60\% to 88.29\%. However, it is unable to fully eliminate the gap, suggesting that prompting alone is insufficient to resolve the cross-modal misalignment.

Post-training with supervised or preference-based objectives further improves performance, yet still falls short of full recovery. SFT and DPO raise the average audio accuracy to 72.52\% and 75.37\%, respectively. This indicates that while supervision and preference optimization help, they do not explicitly align cross-modal reasoning behavior. Standard GRPO, trained with the base reward $R_{\text{base}}$, provides additional improvements (MRR = 92.21\%) but still underperforms DPO, highlighting the limitation of sparse, outcome-centric rewards.
In contrast, our approach achieves the highest performance, reaching \textbf{79.57\%} average audio accuracy and MRR = \textbf{100.28\%}, demonstrating the effectiveness of proposed asymmetric dense alignment reward. Finally, we monitor ASR-related capability via {WER} and observe it remains unchanged ($\approx$4.16\%–4.24\%), supporting the conclusion that gains primarily stem from reasoning alignment rather than improved speech recognition.

\subsection{Reward Components}
\label{sec:ablation}

Table~\ref{tab:rl_effectiveness} presents an ablation study on Phi-4-MM backbone, starting from Standard GRPO trained with the base reward $R_{\text{base}}$ only, and then adding $R_{\text{rep}}$, $R_{\text{beh}}$, or their combination. Incorporating representation alignment reward consistently improves performance, increasing MRR from 92.21\% to 95.56\%, suggesting that aligning layer-wise hidden-state representations provides a denser signal for RL. Alternatively, adding the behavior alignment reward pushes the model close to full recovery (MRR = 99.22\%), indicating that semantic-consistency supervision constrains speech outputs toward correct text-conditioned behaviors. Combining both rewards achieves the best result (MRR = 100.28\%), showing that representation and behavior signals are complementary: representation alignment mitigates representation drift, while behavior alignment enforces semantic target consistency. This validates our objective of jointly aligning internal representations and external behaviors.

\section{Discussion and Analysis}
\label{sec:discussion}

\subsection{Layer-wise Trajectory Alignment}
\label{sec:layer_sensitivity}

\begin{figure*}[h]
    \centering
    \includegraphics[width=\textwidth]{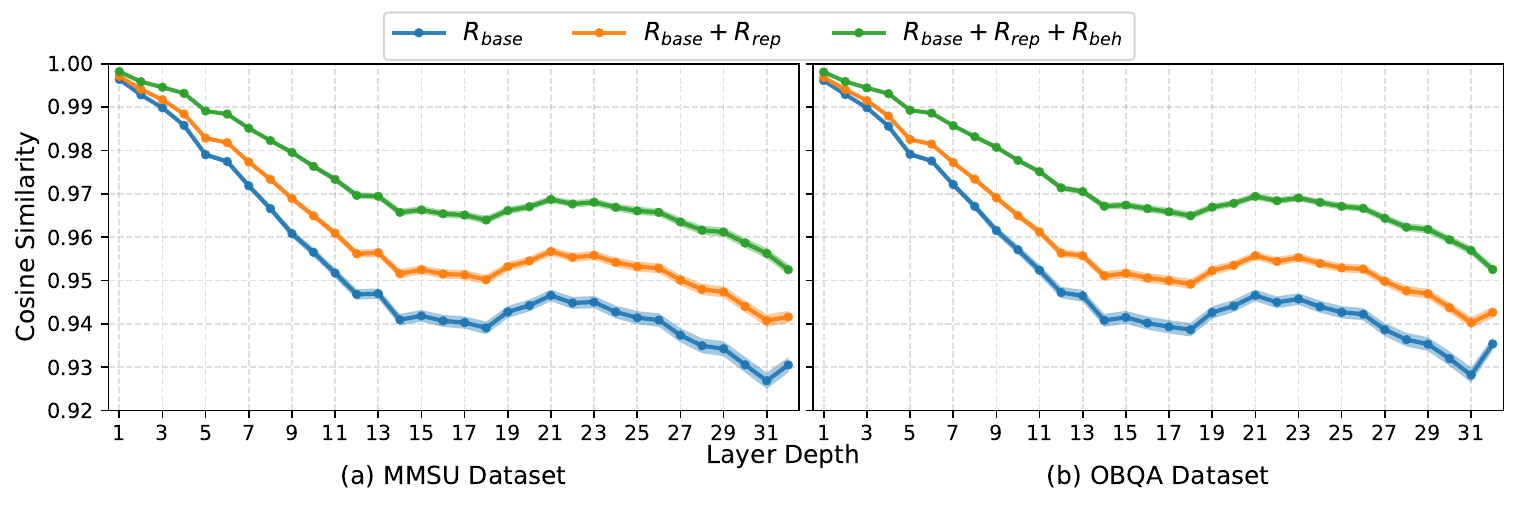}
    \caption{
        {Layer-wise Representation Alignment Analysis.} 
        Shaded areas indicate 95\% confidence intervals.
    }
    \label{fig:layer_wise_analysis}
\end{figure*}

To investigate the inner workings of our trajectory alignment, we analyze the cosine similarity between the internal hidden states of text and speech modalities (detailed in Appendix~\ref{app:layer_wise_analysis}). Specifically, we pass identical queries—one as text and one as audio—through both the baseline and aligned models. For each layer $l$, we compute the average cosine similarity between the sequence of hidden states $\mathbf{H}^{(l)}_{\text{text}}$ and $\mathbf{H}^{(l)}_{\text{speech}}$.

Figure~\ref{fig:layer_wise_analysis} presents a layer-wise representation alignment analysis on the Phi-4-MM backbone, measuring the cosine similarity between the mean hidden states of correct text and speech trajectories across layers. Under this setting, similarity naturally decreases with depth due to compounding transformations, where minor modality-specific differences accumulate as they propagate through the network. The key metric is not the downward trend, but the relative separation between methods across depth. Adding $R_{\text{rep}}$ consistently lifts the similarity trajectory across layers, indicating that representation alignment reduces the representational drift. Furthermore, our joint strategy achieves the highest similarity, suggesting that behavior alignment acts as a complementary constraint that guides the speech branch toward semantically consistent reasoning paths. These internal observations align with the external improvements in MRR, supporting that our method performs genuine reasoning behavior transfer. The consistency of this effect across MMSU and OBQA further validates its robustness.

\subsection{Layer Sensitivity}
\label{sec:layer_sensitivity_ablation}

\begin{figure}[t]
    \centering
    \includegraphics[width=\columnwidth]{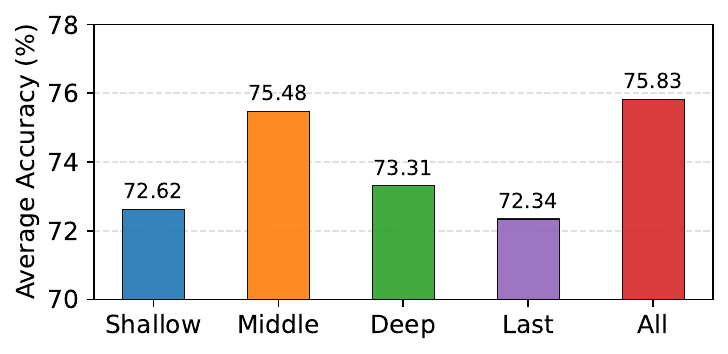}
    \caption{{Sensitivity Analysis of Representation Reward Layers.} Average audio accuracy on MMSU and OBQA across different layer groups.}
    \label{fig:layer_sensitivity}
\end{figure}

Figure~\ref{fig:layer_sensitivity} investigates the sensitivity of the representation-alignment reward to different depths within the 32-layer Phi-4-MM backbone. We partition the model into {Shallow} (layers 1–10), {Middle} (11–20), {Deep} (21–30), and {Last} (31–32) groups, comparing these against an {All}-layer baseline. Results indicate that the {Middle} layers are the most critical localized region, achieving \textbf{75.48\%} accuracy. Conversely, applying the reward exclusively to {Shallow} or {Last} layers is less effective. This suggests that representation drift is primarily in the mid-to-late reasoning stages, rather than during early perceptual processing or final logit alignment. Selecting {All} layers yields the highest accuracy (\textbf{75.83\%}), this confirms that global cross-layer similarity offers advantages over localized alignment.

\subsection{Impact of Alignment Stage on Reward Design}
Ablation on Qwen2.5-Omni (Table~\ref{tab:qwen_ablation}) reveals a stage-dependent regularization effect. As an internal model with minimal post-training (e.g., SFT only), Phi-4-MM is highly receptive to explicit layer-wise trajectory guidance ($R_{\text{rep}}$) to shape its reasoning pathways. In contrast, applying strict hidden-state constraints ($R_{\text{rep}}$) to a highly optimized model like Qwen2.5-Omni causes over-regularization. It forces the speech representations to rigidly mimic text representations at every layer, stripping the model of the necessary internal degrees of freedom to process modality-specific features. This phenomenon is broadly consistent with findings from knowledge distillation for generative models, which suggest that overly rigid imitation objectives can be suboptimal under capacity limitation, while more flexible output-level guidance better accommodates the model's own representational constraints~\cite{gu2024minillm}. Consequently, relying solely on the behavior reward ($R_{\text{beh}}$) yields the optimal guidance for Qwen2.5-Omni, as it enforces semantic consistency at the output while allowing flexible internal reasoning pathways, achieving the highest average audio accuracy (77.71\%) and MRR (100.01\%).

\begin{table}[htbp]
    \centering
    \resizebox{\columnwidth}{!}{
    \begin{tabular}{l|c|c}
    \toprule
    \textbf{Method} & \textbf{Avg. A} & \textbf{MRR (\%)} \\
    \midrule
    Baseline (w/ CoT) & 74.50 & 95.88 \\
    Standard GRPO ($R_{\text{base}}$) & 77.10 & 99.23 \\
    \quad + $R_{\text{rep}}$ & 76.45 & 98.39 \\
    \quad \textbf{+ $R_{\text{beh}}$} & \textbf{77.71} & \textbf{100.01} \\
    \quad + Both ($R_{\text{rep}}$ + $R_{\text{beh}}$) & 76.84 & 98.89 \\
    \bottomrule
    \end{tabular}
    }
    \caption{Ablation study on Qwen2.5-Omni.}
    \label{tab:qwen_ablation}
\end{table}

\subsection{Preservation of Text Reasoning Capability}
\label{sec:text_only_ablation}

To determine whether simultaneous speech alignment compromises text reasoning, we conducted a ``Text-only GRPO'' ablation on the Phi-4-MM backbone, optimizing solely on text inputs with the base reward ($R_{\text{base}}$). As shown in Table~\ref{tab:rl_effectiveness}, Text-only GRPO yields an average text accuracy of 83.76\%, nearly identical to our joint alignment method (83.56\%). This demonstrates that aligning speech trajectories does not interfere with the model's text capabilities. Furthermore, our method achieves higher speech reasoning accuracy (79.57\% vs. 76.86\%) and modality recovery (MRR: 100.28\% vs. 96.86\%). These results confirm that our framework effectively bridges the modality gap while preserving text performance.

\subsection{Generalization to Real Speech and Open-Ended Tasks}
\label{sec:generalization}

To verify that our alignment framework generalizes to real human speech and preserves the model's ability to reason over rich paralinguistic features, we evaluate on two additional benchmarks. We select SD-QA~\cite{sdqa_2021}, a reference-based open-ended QA dataset featuring real spoken questions with diverse accents, and the MMSU benchmark~\cite{mmsu_cuhk}\footnote{Note that this MMSU benchmark differs from the VoiceBench MMSU split used in our main experiments.}, which assesses fine-grained perception of paralinguistic features like prosody. 
As shown in Table~\ref{tab:generalization}, our method consistently outperforms both the Phi-4-MM baseline and standard CoT prompting. Notably, while CoT prompting degrades fine-grained acoustic reasoning (dropping MMSU accuracy to 51.50\%), our alignment framework avoids this regression, achieving 57.40\% on MMSU Audio and 41.95\% on SD-QA Audio. These results demonstrate that aligning reasoning trajectories preserves speech-specific attributes like prosody and accents, rather than overwriting them with text-based patterns, enabling robust generalization to open-ended, real-world spoken interactions.

\begin{table}[htbp]
    \centering
    % \small
    \begin{tabular}{l|c|cc}
    \toprule
    \multirow{2}{*}{\textbf{Method}} & \textbf{MMSU} & \multicolumn{2}{c}{\textbf{SD-QA}} \\
    \cmidrule(lr){2-2} \cmidrule(lr){3-4}
    & \textbf{A} & \textbf{A} & \textbf{T} \\
    \midrule
    Phi-4-MM & 56.70 & 39.42 & 41.77 \\
    + CoT Prompting & 51.50 & 39.60 & \textbf{45.21} \\
    \textbf{\modelname} & \textbf{57.40} & \textbf{41.95} & 44.30 \\
    \bottomrule
    \end{tabular}
    \caption{Generalization performance on paralinguistic perception (MMSU) and real-speech QA (SD-QA).}
    \label{tab:generalization}
\end{table}

\subsection{Domain-Specific Error Analysis}
\label{sec:per_domain_analysis}

To understand how our alignment mitigates representational drift under a free-generation setting, we conducted a per-domain breakdown on the 12 domains of the MMSU dataset (see Appendix~\ref{app:mmsu_domain_table} for full score details). Our method yields consistent improvements across all domains. In high-performing tasks (e.g., Business, Biology, Law), the aligned model demonstrates enhanced arithmetic calculation, logical consistency, and domain knowledge. Conversely, in lower-performing domains (e.g., Economics, Engineering, History), errors often stem from subtle semantic traps or chronological retrieval failures. Detailed qualitative CoT traces and specific case studies are provided in Appendix~\ref{app:qual_analysis}.

\subsection{Computational Efficiency}
\label{sec:computational_efficiency}

Analysis of computational efficiency shows that our online RL approach introduces only a marginal 4.4\% overhead compared to standard GRPO, while achieving better performance (Appendix~\ref{app:efficiency}).

\section{Conclusion}
We introduced an on‑policy trajectory alignment framework that mitigates representational drift and improves semantic consistency in speech reasoning. By combining dense representation and behavior alignment rewards under an asymmetric RL objective, our method substantially narrows the modality reasoning gap and achieves state‑of‑the‑art performance on MMSU and OBQA among 7B‑scale Speech LLMs.

\section*{Limitations}
Despite its effectiveness, our trajectory alignment framework has several limitations. First, we evaluate alignment only at the 7B scale, and it remains unclear how the proposed reward design behaves for {smaller or larger models}. Second, our method focuses on single-turn reasoning, whereas multi-turn, interactive, or dialogue-driven speech reasoning may introduce additional dynamics not captured by our current formulation. Finally, while our alignment rewards mitigate modality drift, they still rely on text-only reference completions and may not fully account for paralinguistic cues, such as emotion, prosody, or intent, that do not have explicit textual counterparts.

% Bibliography entries for the entire Anthology, followed by custom entries
%\bibliography{anthology,custom}
% Custom bibliography entries only
\bibliography{custom}

\appendix

\section{Format Reward Regular Expression}
\label{app:regex}
The format reward, $R_{\text{format}}$, is a binary reward assigned based on a regular expression match. The regular expression is:
\begin{Verbatim}[breaklines=true]
^<think>.*?</think>\s*<answer>.*The answer is [ABCD][\.:,].*</answer>$
\end{Verbatim}

\section{Baseline Details}
\label{app:baseline_details}

We provide detailed descriptions of the baselines compared in our experiments.

\paragraph{Cascaded Systems.}
We construct pipeline baselines using \texttt{whisper-large-v3} as the ASR front-end, paired with Llama3.1-8B, Qwen2.5-7B, and Phi-4-7B (the text backbone $\pi_{\text{base}}$ of Phi-4-MM). For audio evaluation, we transcribe the speech using the ASR model and feed the transcripts into the corresponding text LLM, reflecting the impact of ASR errors on reasoning. For text evaluation, we feed the clean text to the LLM.

\paragraph{Cross-modal Alignment Methods.}
SALAD~\cite{salad} focuses on sample-efficient distillation or targeted data selection to improve speech-text alignment.
DeSTA2.5-Audio~\cite{desta25} utilizes self-generated text completion as the alignment target for cross-modal consistency.
AlignChat~\cite{alignchat} represents a frozen-backbone method, focusing exclusively on the alignment of speech adapters without altering the backbone LLM parameters.
Knowledge Distillation (KD)~\cite{kd} serves as a distillation-based transfer baseline, mimicking the behavior from a text teacher to a speech student.

\paragraph{General Post-training Methods.}
For SFT and DPO, we construct preference data via reject sampling. SFT trains only on the chosen completions, while DPO leverages chosen-rejected pairs. Standard GRPO corresponds to the base-reward-only RL baseline (as described in Section~\ref{sec:framework}), which optimizes task performance without explicitly modeling the modality gap.

\section{Hyperparameters}
\label{app:hyperparams}

Table~\ref{tab:hyperparams} lists the detailed hyperparameters used for the SFT and RL training stages across all experiments. We utilize LoRA for efficient fine-tuning to reduce computational overhead.

\begin{table}[htbp]
    \centering
    \small
    \begin{tabular}{l|c}
    \toprule
    \textbf{Hyperparameter} & \textbf{Value} \\
    \midrule
    \multicolumn{2}{c}{\textit{LoRA Configuration}} \\
    Rank ($r$) & 8 \\
    Alpha ($\alpha$) & 32 \\
    Dropout & 0.05 \\
    Target Modules & All Linear Layers \\
    \midrule
    \multicolumn{2}{c}{\textit{Training Configuration}} \\
    Learning Rate & $1\text{e-}5$ (Qwen), $2\text{e-}5$ (Phi) \\
    LR Scheduler & Cosine \\
    Warmup Ratio & 0.01 \\
    Num Epochs & 3 \\
    Batch Size (Global) & 64 \\
    Gradient Accumulation & 4 \\
    Optimizer & AdamW \\
    Weight Decay & 0.01 \\
    Max Grad Norm & 1.0 \\
    Precision & bfloat16 \\
    \midrule
    \multicolumn{2}{c}{\textit{GRPO / DAPO Configuration}} \\
    Generations per Prompt ($G$) & 8 \\
    Temperature & 1.0 \\
    Max Completion Length & 1024 \\
    Epsilon High ($\epsilon_{\text{high}}$) & 0.28 \\
    KL Coefficient ($\beta$) & 0.0 \\
    \bottomrule
    \end{tabular}
    \caption{Detailed hyperparameters for training.}
    \label{tab:hyperparams}
\end{table}

\section{Prompt Templates}
\label{app:prompts}

We employ same prompt format across all experiments. This instruction enforces a Chain-of-Thought reasoning structure. Table~\ref{tab:prompt_template} illustrates the specific templates used.
The content of \texttt{[QUESTION TEXT]} (and the spoken content of \texttt{<|audio\_1|>}) follows the multiple-choice format:
\begin{Verbatim}
{question}

Option A: {option_a}
Option B: {option_b}
Option C: {option_c}
Option D: {option_d}
\end{Verbatim}

\begin{table}[t!]
    \centering
    \small
    \begin{tabular}{l|p{5cm}}
    \toprule
    \textbf{Component} & \textbf{Content} \\
    \midrule
    \textbf{System} & \texttt{<|system|>A conversation between User and Assistant. The user asks a question, and the Assistant solves it. The assistant first thinks about the reasoning process in the mind and then provides the user with the answer. The reasoning process and answer are enclosed within <think> </think> and <answer> </answer> tags, respectively, i.e., <think>[THINKING PROCESS]</think><answer>The answer is [CHOICE].</answer><|end|>} \\
    \midrule
    \textbf{User (Audio)} & \texttt{<|user|><|audio\_1|><|end|>} \\
    \midrule
    \textbf{User (Text)} & \texttt{<|user|>[QUESTION TEXT]<|end|>} \\
    \midrule
    \textbf{Assistant} & \texttt{<|assistant|>} \\
    \bottomrule
    \end{tabular}
    \caption{Prompt templates for training and inference.}
    \label{tab:prompt_template}
\end{table}

\section{Layer-wise Analysis Methodology}
\label{app:layer_wise_analysis}

For the layer-wise representation alignment analysis presented in Section~\ref{sec:layer_sensitivity}, we perform a teacher-forcing analysis to ensure identical token alignment. We feed the exact same text-conditioned generated CoT response tokens to both the text- and speech-conditioned branches. We align the response span starting from the last \texttt{<|assistant|>} token and compute the cosine similarity of hidden states at each corresponding token position for every layer. The reported curves in Figure~\ref{fig:layer_wise_analysis} are obtained by averaging these similarities over all valid tokens and test samples, accompanied by 95\% confidence intervals, evaluated on both the MMSU and OBQA datasets.

\section{Per-domain Breakdown Analysis}
\label{app:mmsu_domain_table}

Table~\ref{tab:mmsu_domain} presents the detailed per-domain accuracy breakdown on the MMSU dataset, demonstrating that our trajectory alignment method consistently improves performance across all 12 categories.

\begin{table}[htbp]
    \centering
    \small
    \resizebox{\linewidth}{!}{
    \begin{tabular}{l|c|c|c|c}
    \toprule
    \textbf{Category} & \textbf{Count} & \textbf{Baseline (w/ CoT)} & \textbf{\modelname} & \textbf{Delta} \\
    \midrule
    Business & 236 & 61.02\% & 75.00\% & +13.98\% \\
    Biology & 172 & 69.77\% & 83.72\% & +13.95\% \\
    Law & 51 & 52.94\% & 66.67\% & +13.73\% \\
    Physics & 383 & 61.36\% & 73.89\% & +12.53\% \\
    Health & 406 & 56.65\% & 66.75\% & +10.10\% \\
    Philosophy & 305 & 46.89\% & 55.08\% & +8.20\% \\
    Chemistry & 167 & 65.87\% & 73.05\% & +7.19\% \\
    Psychology & 317 & 64.04\% & 70.66\% & +6.62\% \\
    Economics & 280 & 74.64\% & 80.71\% & +6.07\% \\
    Other & 546 & 60.99\% & 66.85\% & +5.86\% \\
    History & 104 & 51.92\% & 57.69\% & +5.77\% \\
    Engineering & 107 & 56.07\% & 61.68\% & +5.61\% \\
    \bottomrule
    \end{tabular}
    }
    \caption{Per-domain breakdown on MMSU Audio Accuracy.}
    \label{tab:mmsu_domain}
\end{table}

\section{Training Efficiency}
\label{app:efficiency}

We provide a runtime and memory analysis of our method compared to offline strategies on 8$\times$H200 GPUs for 3 epochs. As shown in Table~\ref{tab:efficiency}, standard SFT and DPO require approximately 1.6 hours for training. However, this excludes the substantial offline data generation cost, which exceeds 10.9 GPU-hours to generate 8 rollouts per prompt. In contrast, our method operates entirely online. Compared to the direct Standard GRPO baseline (27.5 hours), computing our alignment rewards introduces only a marginal computational overhead of 4.4\%, totaling 28.7 hours. Furthermore, our method maintains an identical peak memory footprint of 90\% relative to the GRPO baseline.

\begin{table}[htbp]
    \centering
    \small
    \begin{tabular}{l|c|c|c}
    \toprule
    \multirow{2}{*}{\textbf{Method}} & \textbf{Peak} & \textbf{Data Gen.} & \textbf{Training} \\
    & \textbf{Memory} & \textbf{Time} & \textbf{Time} \\
    \midrule
    SFT & 98\% & 10.9h$^*$ & 1.6h \\
    DPO & 95\% & 10.9h$^*$ & 1.6h \\
    Standard GRPO & 90\% & N/A & 27.5h \\
    \textbf{\modelname} & 90\% & N/A & 28.7h \\
    \bottomrule
    \end{tabular}
    \caption{Efficiency comparison. $^*$Excludes additional time required for reward scoring.}
    \label{tab:efficiency}
\end{table}

\section{Qualitative Analysis}
\label{app:qual_analysis}

\begin{figure}[h!]
    \centering
    \includegraphics[width=\columnwidth]{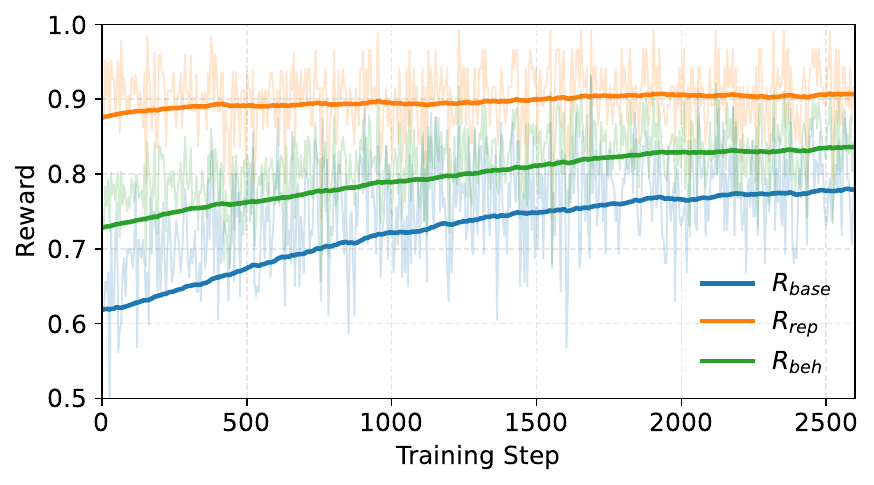}
    \caption{Reward curves during training.}
    \label{fig:reward_curve}
\end{figure}

\paragraph{Free-Generation Alignment Analysis.} 
To investigate how our method performs under a free-generation setting without teacher-forcing, we analyze the reward curves during the online RL training process. Since the training rollouts are generated autoregressively by the policy, these curves intrinsically reflect the model's performance under free generation. As shown in Figure~\ref{fig:reward_curve}, the upward trends of the base reward ($R_{\text{base}}$), representation alignment reward ($R_{\text{rep}}$), and behavior alignment reward ($R_{\text{beh}}$) demonstrate consistent improvements in base task accuracy, as well as in the representation similarity and semantic consistency between speech and text modalities. 

\paragraph{Training Improvements.} Table \ref{tab:qual_analysis} shows specific qualitative improvements. The trained model exhibits better math calculation, logical reasoning, and domain knowledge, successfully correcting previous hallucinations and reasoning failures. 

\paragraph{Persistent Errors.} Table \ref{tab:error_analysis} analyzes cases where the model fails to reach the ground truth after training, reflecting the challenges in lower-performing domains. This reveals that failures often stem from subtle semantic traps, conflicting domain logic, or chronological retrieval failures.

\begin{table*}[t!]
\centering
\small
\renewcommand{\arraystretch}{1.5}
\resizebox{\textwidth}{!}{
\begin{tabular}{p{0.18\linewidth} p{0.32\linewidth} p{0.32\linewidth} p{0.18\linewidth}}
\toprule
\textbf{Question} & \textbf{Before Training} & \textbf{After Training} & \textbf{Analysis} \\
\midrule

% Row 1: Business/Math
\textbf{[Business]} Paul took a loan of \$384.75 for 60 days at 6\% interest. How much interest did he pay? (GT: B. \$3.85) & 
<think> ...we need to convert the time to years... plug in the values and solve for the interest. </think> \newline
<answer> The answer is \underline{(A) \$5.75}. </answer> 
& 
<think> ...$I = \$384.75 \times 0.06 \times (60/365)$... $I \approx \$3.79$... The closest option to our calculated interest of \$3.79 is \underline{option B}, which is \underline{\$3.85}. </think> \newline
<answer> The answer is \underline{B: \$3.85}. </answer> 
& 
\textbf{Math Calculation:}
Model executed detailed arithmetic steps and correctly matched the closest option despite rounding variations. \\
\midrule

% Row 2: Biology/Logic
\textbf{[Biology]} Which of the following about meiosis is \underline{NOT} true? (GT: C. Sister chromatids separate during meiosis 1.) & 
<think> ...Option C says sister chromatids separate during meiosis I. \underline{This is not true}... I will choose \underline{option A} as the final answer, because it is the most straightforward mistake. </think> \newline
<answer> \underline{A} </answer> 
& 
<think> ...C. Sister chromatids separate during meiosis I - \underline{This is not true}. Sister chromatids do not separate during meiosis I; instead, homologous chromosomes separate... </think> \newline
<answer> The answer is \underline{C}. </answer> 
& 
\textbf{Logical Reasoning:}
Model resolved the logical conflict in ``NOT/EXCEPT'' questions, consistently linking its evaluation to the final answer. \\
\midrule

% Row 3: Law/Knowledge
\textbf{[Law]} Hume's attack on natural law is founded on his argument that: (GT: A. We cannot objectively know what is morally right or wrong.) & 
<think> ...Hume was a skeptic... did not focus on protection... best answer reflects his skepticism... </think> \newline
<answer> \underline{B. Natural law is backward-looking.} </answer> 
& 
<think> ...He questioned the idea that there are \underline{objective moral truths} that can be known a priori... representation of Hume's attack... is his argument that we \underline{cannot objectively know} what is \underline{morally right or wrong}... </think> \newline
<answer> The answer is \underline{A}. </answer> 
& 
\textbf{Domain Knowledge:}
Model corrected its domain knowledge, replacing a generic guess with the specific epistemological argument of the philosopher. \\

\bottomrule
\end{tabular}
}
\caption{{Qualitative analysis of training improvement. Bold text denotes the question category. GT denotes Ground Truth. Underlined text highlights improvements.}}
\label{tab:qual_analysis}
\end{table*}

\begin{table*}[h!]
\centering
\small
\renewcommand{\arraystretch}{1.5}
\resizebox{\textwidth}{!}{
\begin{tabular}{p{0.18\linewidth} p{0.29\linewidth} p{0.29\linewidth} p{0.24\linewidth}}
\toprule
\textbf{Question} & \textbf{Before Training} & \textbf{After Training} & \textbf{Analysis} \\
\midrule

% Row 1: Economics (Logical Trap / Definition Error)
\textbf{[Economics]} If the reserve ratio is 10 percent and a new customer deposits \$500, what is the maximum amount of money created? (GT: A. \$4,500) & 
<think> ... so the money multiplier is $1 / 0.10 = 10$. The maximum amount of money created is the initial deposit multiplied by the multiplier... \underline{$\$500 \times 10 = \$5000$}. </think> \newline
<answer> \underline{B) \$5000} </answer> 
& 
<think> ...the money multiplier would be 1 divided by 0.10 ... equals 10. ...the maximum amount of money that can be created is the initial deposit multiplied by the money multiplier... \underline{\$500 ... * 10 ... = \$5000}... </think> \newline
<answer> The answer is \underline{B: \$5000}. </answer> 
& 
\textbf{Logical Trap:} It correctly calculates the \textit{total} money supply: \$5000, but fails by missing the nuance: ``new money \textit{created}'', which requires subtracting the deposit: \$5000 - \$500 = \$4500. \\

\midrule

% Row 2: Engineering
\textbf{[Engineering]} The resistance in the circuit of the moving coil of a dynamometer wattmeter should be: (GT: C. High) & 
<think>...\underline{the moving coil is used to} \underline{measure the current}... The resistance \underline{should be low to minimize power loss}... </think> \newline
<answer>\underline{(B) very low}</answer> & 
<think> ...The moving coil should have a very low resistance ... the power consumed ... If the resistance were high, it would consume more power... \underline{A resistance that is almost zero} \underline{would be ideal}... </think> \newline
<answer> The answer is \underline{B: very low}. </answer> & 
\textbf{Logical Error:} The model wrongly assumes low resistance prevents power consumption. Because the moving (voltage) coil is connected in parallel, $P = V^2/R$ governs; thus, a \textit{low} resistance would actually \textit{increase} power draw and cause a short. \\

\midrule

% Row 3: History (Semantic/Knowledge Conflict)
\textbf{[History]} Which of the following is the last to be occupied by human beings? (GT: D. New Zealand) & 
<think> ...Hawaii: ...arrived in Hawaii around \underline{1,500 years ago}... New Zealand: ...arrived in New Zealand around \underline{1,000 years ago}... the \underline{last place} to be occupied... \underline{is Hawaii}. </think> \newline
<answer> \underline{C} </answer> 
& 
<think> ...\underline{Hawaii} ... settled ... \underline{around 300-500 CE}... \underline{New Zealand} ... settled ... \underline{around 1300 CE}, which is ... \underline{earlier than Hawaii}. Comparing these dates... \underline{Hawaii was the last} to be settled... </think> \newline
<answer> The answer is \underline{C: Hawaii}. </answer> 
& 
\textbf{Chronological Error:} The model retrieves the correct historical dates (New Zealand at 1300 CE vs. Hawaii at 300-500 CE) but spectacularly fails at basic temporal comparison, wrongly deducing that 1300 CE is ``earlier'' than 300 CE. \\

\bottomrule
\end{tabular}
}
\caption{{Qualitative analysis of persistent errors. GT denotes Ground Truth. Underlined text highlights incorrect reasoning steps.}}
\label{tab:error_analysis}
\end{table*}

\end{document}